\pdfoutput=1
\documentclass[11pt]{article}

\usepackage[final]{acl}

\usepackage{times}
\usepackage{latexsym}
\usepackage{amssymb}
\usepackage{amsmath}
\usepackage{booktabs}
\usepackage{multirow}

\usepackage[T1]{fontenc}
\usepackage[utf8]{inputenc}

\usepackage{microtype}

\usepackage{inconsolata}

\usepackage{graphicx}
\usepackage{cancel}
\title{MMUTF: Multimodal Multimedia Event Argument Extraction with Unified Template Filling}

\author{
 \textbf{Philipp Seeberger}, 
 \textbf{Dominik Wagner},
 \textbf{Korbinian Riedhammer}
\\
 Technische Hochschule Nürnberg Georg Simon Ohm \\
  \small$\texttt{\{philipp.seeberger, dominik.wagner, korbinian.riedhammer\}@th-nuernberg.de}$
}

\begin{document}
\maketitle

% \linespread{0.965}\selectfont

\begin{abstract}
With the advancement of multimedia technologies, news documents and user-generated content are often represented as multiple modalities, making Multimedia Event Extraction (MEE) an increasingly important challenge.
However, recent MEE methods employ weak alignment strategies and data augmentation with simple classification models, which ignore the capabilities of natural language-formulated event templates for the challenging Event Argument Extraction (EAE) task.
In this work, we focus on EAE and address this issue by introducing a unified template filling model that connects the textual and visual modalities via textual prompts.
This approach enables the exploitation of cross-ontology transfer and the incorporation of event-specific semantics.
Experiments on the M2E2 benchmark demonstrate the effectiveness of our approach. 
Our system surpasses the current SOTA on textual EAE by +7\% F1, and performs generally better than the second-best systems for multimedia EAE.
\end{abstract}

\section{Introduction}\label{sec:introduction}

\begin{figure}[h]
  \centering
\includegraphics[width=1.0\linewidth]{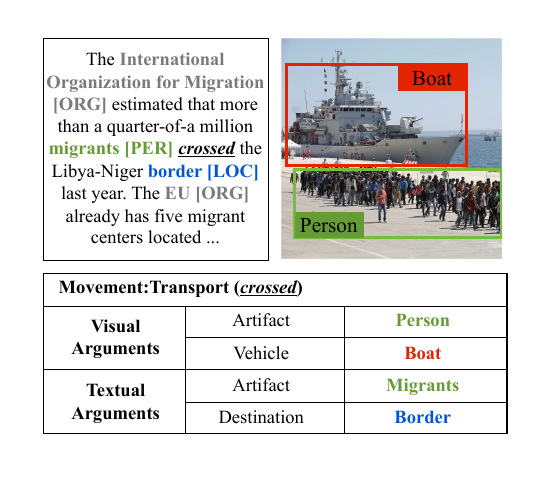}
  \vspace{-8mm}
  \caption{Example of Multimedia Event Extraction from the M2E2 benchmark, where event argument roles are extracted from both the textual and visual modality.}
  \label{fig:example}
\end{figure}

Traditional Event Extraction (EE) consists of two tasks: Event Detection (ED) and Event Argument Extraction (EAE), which address event type identification and argument role extraction, respectively.
Traditional methods focus on individual modalities, such as text, images, or videos \citep{huang_textee_2023,yatskar_situation_2016,khan_grounded_2022}. 
However, focusing on single modalities results in incomplete event understanding, as up to 33\% of visual arguments are absent in the accompanying text \citep{li_cross-media_2020} (cf.~\autoref{fig:example}).

As the creation of multimodal EE datasets is time-consuming and costly, recent advanced MEE methods focus on (weak) image-text alignment based on available unimodal task-dependent and task-independent vision-language datasets \citep{li_cross-media_2020,liu_multimedia_2022,du_training_2023,liu_multi-grained_2024}.
However, current multimedia EAE models are still based on simple classification techniques while ignoring cross-ontology transfer capabilities and event template semantics.

To address this gap, we reformulate existing query-based (template filling) methods \citep{wang_query_2022,zhang_transfer_2022,ma_prompt_2022,zheng_query_2023} into a unified template filling framework, which enables us to utilize event templates as natural language prompts for different input modalities.
First, our \textbf{M}ultimodal \textbf{M}ultimedia Event Argument Extraction with \textbf{U}nified \textbf{T}emplate \textbf{F}illing (MMUTF) model exploits candidate structures, such as textual entities and visual objects, and connects them via query representations (i.e., argument roles) in a unified latent space.
Finally, these representations are used to match event argument roles with the corresponding candidates via queries extracted from event templates.

\paragraph{Contributions} We summarize our main contributions as follows: 1) We introduce MMUTF, a unified template filling framework that addresses the EAE task for multiple input modalities. 2) We validate our approach on the widely used M2E2 benchmark and demonstrate superior performance compared to the majority of baselines.
3) We analyse the transfer learning capabilities for multimedia EAE by employing FrameNet as a resource-rich data source.
\section{Method}\label{sec:method}

\begin{figure*}[h]
  \centering
\includegraphics[width=1.0\linewidth]{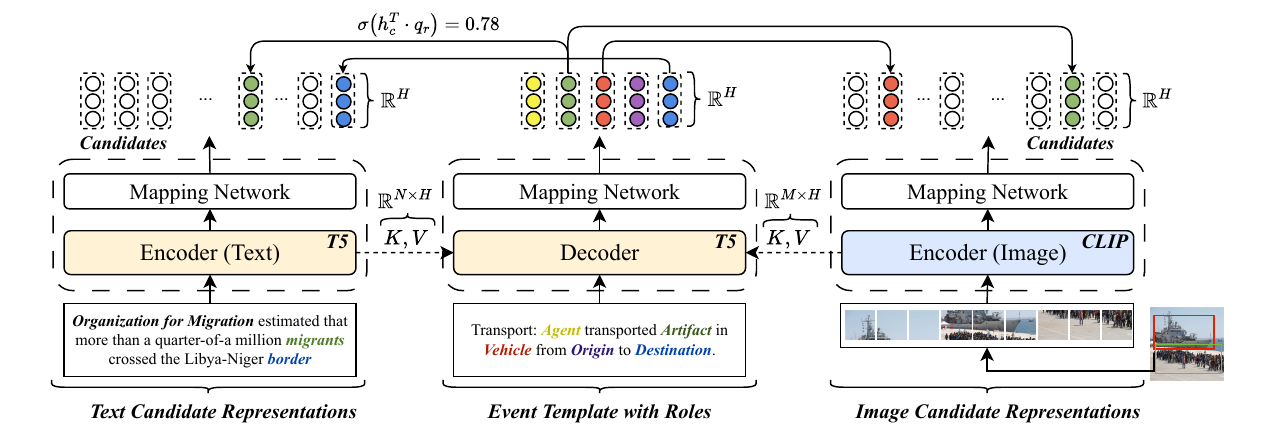}
  \caption{The overall architecture of \textsc{MMUTF}. Given a text or image event, our model encodes the corresponding event template and argument roles as a textual prompt. The textual or image context interacts with the prompt via the cross-attention mechanism, resulting into candidate and query representation vectors. Here, the candidates correspond to entities or objects, depending on the modality. A matching score then assigns the candidates to the argument roles based on a predefined threshold.}
  \label{fig:approach}
\end{figure*}

For event argument roles, the extraction task can be framed as a candidate-query matching problem, where argument roles serve as the queries.
We can utilize the matching between candidates and queries to address the multimedia EAE task with a unified template filling model.
Specifically, we aim to create candidate representations for either textual entities or image objects and compute a matching score for each query (i.e., argument role) in a given event template.\footnote{We can also consider text spans, image segments, etc., as candidates. However, entities and objects follow the present annotation schemes and suit the EAE task best.}

\subsection{Multimodal Unified Template Filling}

Our unified template filling model is depicted in \autoref{fig:approach}.
We use both the sentences and images from a multimedia document as input contexts, with the written descriptions of event templates and argument roles acting as placeholders.
We approach the EAE task in multiple steps: 1) Given a predicted event trigger, we select the corresponding event template. 2) We encode the input sentences and images with modality-specific encoder models and compute candidate representations. 3) We feed the event template into the query model and obtain query representations for each role. 4) Lastly, we compute the matching scores and treat the EAE task as a binary classification problem.

\paragraph{Candidate Representations} In order to obtain modality-specific candidates, we recognize entities and objects with off-the-shelf systems and extract contextualized representations using a Transformer model for each modality \citep{vaswani_attention_2017,raffel_exploring_2020,radford_learning_2021}.
The encoding process yields a sequence of $N$ token embeddings $t=\{t_{i}\}_{1}^{N} \in \mathbb{R}^{N \times H}$ and $M$ patch embeddings $p=\{p_{i}\}_{1}^{M} \in \mathbb{R}^{M \times H}$ with dimensionality $H$ for each input sentence $S$ and image $I$, respectively.
For entities tokenized into multiple subwords, we use mean pooling and concatenate the feature vector with the representation of the event trigger word.
Similarly, for images, we obtain object feature vectors using the max pooling\footnote{Max pooling achieved the best results among other techniques such as mean and RoI pooling.} operation over all patches containing the detected object, and concatenate these with the image \textit{\textsc{CLS}} embeddings for further processing.
Here, the \textit{\textsc{CLS}} embeddings serve as event triggers represented as image scenes.
The concatenated vectors are then transformed via their corresponding mapping networks, which will be described in more detail later.

\paragraph{Query Representations} For event template processing, we employ the decoder of a Transformer Encoder-Decoder model \citep{raffel_exploring_2020} as shown in \autoref{fig:approach}.
This approach has shown promising results without compromising on the length of the input context, as seen in encoder architectures \citep{ma_prompt_2022}. 
This configuration enables the prompt to interact with the input contexts at the cross-attention layers of every decoder block, resulting in context-dependent query representations.
Similar to the generation of candidate representations, we use the mean pooling operation for argument roles that span multiple subwords and feed the features into a subsequent mapping network.
We denote the query representations as $q=\{q_{i}\}_{1}^{R} \in \mathbb{R}^{R \times H}$, where $H$ is the dimensionality, and $R$ is the number of argument roles. 

\paragraph{Mapping Network} We aim to align the modality-specific candidate and query representations into a unified latent space, based on the candidate and query feature vectors.
However, during training, we face the challenge of unifying the rich and diverse representations of the different modalities and Transformer models. 
The mapping networks are designed to bridge the gap between the latent spaces and transform all candidate and query features into ${H}$-sized representations.
Each mapping network shares the same architecture and consists of a two-layer feedforward network with $4 \times H$ hidden units, ReLU \citep{relu_2018} activation, and is trained with a dropout probability of 40\%.

\subsection{Training and Inference}

For each event, we evaluate whether a candidate $c$ represents an argument role $r$ by computing a matching score $\phi(c,r) = \sigma (\textbf{h}_{c}^{T}\textbf{q}_{r}) \in \mathbb{R}$.
Here, $\textbf{h}_{c}$ denotes the representation vector of the candidate $c$, $\textbf{q}_{r}$ indicates the query vector of argument role $r$, and $\sigma$ represents the sigmoid activation.
The model parameters and representations are optimized to enhance the matching scores for positive and to reduce it for negative candidate-query pairs:
\begin{equation}
    \mathcal{L_{\text{BCE}}} = - \frac{1}{|C|} \sum_{i=1}^{|C|} \sum_{j=1}^{|A|} y_{ij} \ log \, \phi(i,j)
\end{equation}
where $\mathcal{L_{\text{BCE}}}$ denotes the binary cross-entropy loss, $y_{ij}$ the ground truth label, $C$ the collection of candidates and $A$ the set of possible argument roles.
At inference time, we use a threshold $\tau$ with $\phi(c,r) \ge \tau$ to decide whether candidate $c$ corresponds to argument role $r$.
For each event and candidate, we assign the role with the highest score that exceeds the threshold $\tau$.
Otherwise, the candidate is assigned to the \textit{None} class.
\section{Experiments}\label{sec:experiments}

\begin{table*}[t]
    \centering
    \resizebox{\linewidth}{!}{
    \begin{tabular}{lccc|ccc|ccc|ccc|ccc|ccc}
        \toprule
        \multicolumn{1}{c}{} & \multicolumn{6}{c}{\textbf{Textual Events}} & \multicolumn{6}{c}{\textbf{Visual Events}} & \multicolumn{6}{c}{\textbf{Multimedia Events}} \\
        & \multicolumn{3}{c}{\textbf{Event Mention}} & \multicolumn{3}{c}{\textbf{Argument Role}} & \multicolumn{3}{c}{\textbf{Event Mention}} & \multicolumn{3}{c}{\textbf{Argument Role}} & \multicolumn{3}{c}{\textbf{Event Mention}} & \multicolumn{3}{c}{\textbf{Argument Role}} \\
        \textbf{Model} & P & R & F1 & P & R & F1 & P & R & F1 & P & R & F1 & P & R & F1 & P & R & F1 \\
        \toprule
        \textsc{FLAT\textsubscript{Att}} & 34.2 & 63.2 & 44.4 & 20.1 & 27.1 & 23.1 & 27.1 & 57.3 & 36.7 & 4.3 & 8.9 & 5.8 & 33.9 & 59.8 & 42.2 & 12.9 & 17.6 & 14.9 \\
        \textsc{FLAT\textsubscript{Obj}} & 38.3 & 57.9 & 46.1 & 21.8 & 26.6 & 24.0 & 26.4 & 55.8 & 35.8 & 9.1 & 6.5 & 7.6 & 34.1 & 56.4 & 42.5 & 16.3 & 15.9 & 16.1\\
        \textsc{WASE\textsubscript{Att}} & 37.6 & 66.8 & 48.1 & 27.5 & 33.2 & 30.1 & 32.3 & 63.4 & 42.8 & 9.7 & 11.1 & 10.3 & 38.2 & 67.1 & 49.1 & 18.6 & 21.6 & 19.9 \\
        \textsc{WASE\textsubscript{Obj}} & 42.8 & 61.9 & 50.6 & 23.5 & 30.3 & 26.4 & 43.1 & 59.2 & 49.9 & 14.5 & 10.1 & 11.9 & 43.0 & 62.1 & 50.8 & 19.5 & 18.9 & 19.2 \\
        \textsc{CLIP\textsubscript{Event}} & - & - & - & - & - & - & 41.3 & 72.8 & 52.7 & 21.1 & 13.1 & 17.1 & - & - & - & - & - & - \\
        \textsc{UniCL}& 49.1 & 59.2 & 53.7 & 27.8 & 34.3 & 30.7 & 54.6 & 60.9 & 57.6 & 16.9 & 13.8 & 15.2 & 44.1 & 67.7 & 53.4 & 24.3 & 22.6 & 23.4 \\
        \textsc{MGIM} & 50.1 & 66.5 & 55.8 & 28.2 & 34.7 & \underline{31.2} & 55.7 & 64.4 & 58.5 & 24.1 & 14.1 & 17.8 & 46.3 & 69.6 & 55.6 & 25.2 & 21.7 & 24.6 \\
        \textit{\textsc{UMIE}}\textsuperscript{*} & - & - & - & - & - &  - & - & - & - & - & - &  - & - & - & \textit{62.1} & - & - & \textit{24.5} \\
        \textsc{CAMEL} & 45.1 & 71.8 & 55.4 & 24.8 & 41.8 & 31.1 & 52.1 & 66.8 & 58.5 & 21.4 & 28.4 & \textbf{24.4} & 55.6 & 59.5 & 57.5 & 31.4 & 35.1 & \textbf{33.2} \\
        \textsc{CAMEL\textsuperscript{$\dagger$}} & \textit{48.5} & \textit{65.0} & \textit{55.5} & \textit{30.3} & \textit{33.6} & \textit{31.9}  & \textit{55.1} & \textit{59.1} & \textit{57.0} & \textit{21.7} & \textit{22.1} & \textit{21.9} & \textit{47.9} & \textit{63.4} & \textit{54.6} & \textit{25.9} & \textit{30.4} & \textit{27.9} \\
        \textbf{\textsc{MMUTF}} & 48.5 & 65.0 & 55.5 & 33.6 & 44.2 & \textbf{38.2} & 55.1 & 59.1 & 57.0 & 23.6 & 18.8 & \underline{20.9} & 47.9 & 63.4 & 54.6 & 39.9 & 20.8 & \underline{27.4} \\
        \bottomrule
    \end{tabular}}
    \caption{Comparison with SOTA methods on the M2E2 benchmark for textual, visual, and multimedia events. \textbf{Bold} numbers indicate the best EAE model whereas \underline{underlined} metrics denote the second best. $\dagger$ denotes our reproduced results and * is not directly comparable due to lack of evaluation details.}
    \label{table:results:main}
\end{table*}

\paragraph{Datasets} We conduct our experiments on the widely used M2E2 multmedia event extraction benchmark \citep{li_cross-media_2020} with 8 event types and 15 argument roles.
The dataset contains 245 multimedia documents with 6,167 sentences and 1,014 images. 
It encompasses 1,297 textual events and 391 visual events, with 192 textual events and 203 visual events aligning to 309 multimedia events (cf.~\autoref{fig:example}).
Since the M2E2 dataset does not provide a training split, we follow previous work \citep{li_cross-media_2020,liu_multimedia_2022,du_training_2023} and employ ACE2005 and SWiG for the training phase.
ACE2005 \citep{walker_christopher_ace_2006} consists of 33 textual event types and 36 argument roles.
SWiG \citep{vedaldi_grounded_2020} annotates visual events with 504 activity verbs and 1,788 semantic roles.
Following \citet{li_cross-media_2020}, we align both datasets with the M2E2 event ontology.

\paragraph{Metrics} For evaluation, we follow previous studies on EAE \citep{huang_textee_2023} and use Precision (P), Recall (R), and F1-Score (F1) as our evaluation metrics. 
In this work, we focus on the more challenging EAE task but also report event classification results for completeness.

\paragraph{Baselines} We compare our proposed approach with a wide range of state-of-the-art models.
These models include \textsc{FLAT} \citep{li_cross-media_2020}, \textsc{WASE} \citep{li_cross-media_2020}, \textsc{CLIP\textsubscript{Event}} \citep{li_clip-event_2022}, \textsc{UniCL} \citep{liu_multimedia_2022}, \textsc{CAMEL} \citep{du_training_2023}, \textsc{MGIM} \citep{liu_multi-grained_2024}, and \textsc{UMIE} \citep{sun_umie_2024}.
The majority of baselines set the focus on weak image-text alignment but neglect more sophisticated classification approaches w.r.t. EAE.
We provide a more comprehensive overview in \ref{sec:appendix:baselines}.

\paragraph{Experimental Setup} To ensure a fair comparison with the baselines, we utilize the CLIP-base model \citep{radford_learning_2021} as visual encoder with a $16\times16$ patch size, and the T5-base \citep{raffel_exploring_2020} model as textual encoder-decoder.
We generate object candidates with YOLOv8 \citep{jocher_ultralytics_2023} trained on the COCO \citep{fleet_microsoft_2014} dataset and use entities as textual span candidates.

Our system is trained jointly for 5 epochs on the visual task and 25 epochs on the textual task with a batch size of 16 and 32, respectively.
The varying number of epochs reduces the imbalance of modalities in the training datasets.
For optimization, we use AdamW \citep{loshchilov2018decoupled} with an initial learning rate of $3 \times 10^{-5}$, weight decay of $1 \times 10^{-3}$, and a linear scheduler.
We select the best model checkpoint based on F1 of the ACE2005 test dataset and set the default threshold $\tau = 0.5$ for inference.
We report threshold tuning and implementation details in \ref{sec:appendix:threshold} and \ref{sec:appendix:implementation}.

Since \textsc{MMUTF} focuses on EAE and requires event predictions for a realistic comparison, we ran the CAMEL event detection model to obtain event predictions.
We achieve comparable results with a F1 of 55.5\%, 57.0\%, and 54.6\% for textual, visual, and multimedia event detection, respectively. 
However, we observe an absolute performance drop of 1.5\% and 2.9\% for the visual and multimedia event detection tasks. 

\subsection{Results}\label{sec:results}

\begin{table}[t]
    \centering
    \resizebox{\linewidth}{!}{
    \begin{tabular}{lccc}
        \toprule
         \textbf{Strategy} & \textbf{Textual F1} & \textbf{Visual F1} & \textbf{MM F1} \\
        \toprule
        \textsc{Text + Image} & 38.2 & 20.9 & 27.4 \\
        \textsc{Text $\rightarrow$ Image} & 38.2 & 19.0 & 26.9 \\
        \textsc{Image $\rightarrow$ Text} & 36.6 & 20.4 & 26.5 \\
        \textsc{Image Locked} & 38.7 & 20.1 & 27.7 \\
        \bottomrule
    \end{tabular}}
    \caption{Comparison of multimodal training strategies.}
    \label{table:results:training}
\end{table}

The results of our main experiments can be found in \autoref{table:results:main}.
We find that our model surpasses the SOTA in terms of textual EAE by +7\% F1. 
For visual and multimedia EAE, we generally perform better than the second-best systems, achieving improvements of +3.1\% and +2.8\% F1, respectively.
These results demonstrate the efficacy of our approach by utilizing cross-ontology transfer capabilities and event semantics via natural language prompts.
However, we observe a gap of -5.8\% F1 in multimedia EAE compared to \textsc{CAMEL}, primarily due to the visual modality.
With the same set of predicted event mentions, the gap narrows to -0.5\% F1 in multimedia EAE.
We hypothesize that \textsc{CAMEL}'s better visual performance is related to its additional data augmentation pipeline and the integration of textual clues.
Note that we do not utilize any additional image-language paired datasets or cross-modal data augmentation in this work.
Incorporating cross-modal information from the document's context and external datasets, as well as enhanced event detection, might further improve our model's performance.

\subsection{Analysis}\label{sec:analysis}

\paragraph{Multimodal Training Strategy} \autoref{table:results:training} presents results for different multimodal training strategies, focusing on cross-modality alignment for EAE.
\textsc{Text + Image} denotes the training strategy that jointly trains the textual and visual input modalities.
\textsc{Text $\rightarrow$ Image} represents the sequential procedure.
Here, we first train our model on the text input modality, then freeze the text and query model parameters before training the visual input modality.
With \textsc{Image $\rightarrow$ Text}, we follow the same procedure as before, but in reverse.
Lastly, we follow recommendations in vision-language alignment \citep{zhai_lit_2022} and freeze the visual encoder during joint training (\textsc{Image Locked}).
The results suggest that joint training achieves superior performance, while the sequential procedure experiences performance drops of up to -1.9\% F1 for the aligned modality. 
The \textsc{Image Locked} approach yields the best textual and multimodal F1 scores, while also representing the most memory-efficient alternative during training.

\paragraph{Ablation Study}

To better illustrate the effectiveness of different components, we conduct ablation studies and present the results in \autoref{table:results:ablations}.
In line 1, we remove the cross-attention mechanism between the candidate and query representations.
This results in a performance reduction for visual and multimedia EAE, while we observe a slight increase for textual EAE.
In line 2, we replace the joint model for query representations with two modality-specific models.
The performance of all EAE cases significantly declines by up to +3.4\% in terms of F1 scores.
These observations suggest the effectiveness of unified template filling, particularly for the visual and multimedia EAE tasks.
Lastly, to assess the influence of natural language-formulated event templates, we replace the textual query representations with trainable argument role prototypes in line 3.
The performance declines mostly for visual and multimedia EAE, indicating that event template semantics provide beneficial information to enhance cross-ontology transfer capabilities.
We provide backbone and prompt template ablations in \ref{sec:appendix:backbones} and \ref{sec:appendix:prompts}, respectively.

\begin{table}[t]
    \centering
    \resizebox{\linewidth}{!}{
    \begin{tabular}{lccc}
        \toprule
        \textbf{Model} & \textbf{Textual F1} & \textbf{Visual F1} & \textbf{MM F1} \\
        \toprule
        \textsc{MMUTF} & 38.2 & 20.9 & 27.4 \\
        1 \hspace{0.3em} w/o Cross-Attention & 38.5 & 17.8 & 26.5 \\
        2 \hspace{0.3em} w/o Joint-Prompts & 37.0 & 17.5 & 24.4 \\
        3 \hspace{0.3em} w/o Prompts & 37.9 & 15.9 & 23.3 \\
        \bottomrule
    \end{tabular}}
    \caption{Ablations for the different components. ($1$) without cross-attention between candidate and query representations. ($2$) Replacement of joint query model with modality-specific models. ($3$) Replacement of textual query representations with trainable prototypes.}
    \label{table:results:ablations}
\end{table}

\paragraph{Transfer Learning from SRL} Previous work has shown that semantic role labeling (SRL) can serve as a valuable resource for EAE \citep{zhang_transfer_2022}.
Analogous task similarities are also present in image activity recognition, which utilizes FrameNet verbs and frame elements as semantic roles \citep{yatskar_situation_2016}.
However, no work has yet explored the transfer learning capabilities for multimedia EAE.
We use FrameNet 1.7 \citep{baker_berkeley_1998} and SWiG \citep{vedaldi_grounded_2020} as our training datasets and merge their ontologies to 155 frames.
Due to the large number of frames, the event templates are constructed with a semi-automatic approach \citep{zhang_transfer_2022} (cf. \ref{sec:appendix:templates}).
\autoref{table:results:framenet} shows the results for zero-shot prediction on the M2E2 benchmark with gold and predicted event triggers.
We observe remarkable transfer learning capabilities without any training on the M2E2 ontology.
In fact, this simplistic approach outperforms even the supervised baselines FLAT and WASE by up to +11.6\% and +5.8\% F1 for visual and multimedia EAE, respectively.

\begin{table}[t]
    \centering
    \resizebox{\linewidth}{!}{
    \begin{tabular}{lccc}
        \toprule
         \textbf{Model} & \textbf{Textual F1} & \textbf{Visual F1} & \textbf{MM F1} \\
        \toprule
        \multicolumn{4}{l}{\textbf{\textit{Gold Triggers}}} \\
        \textsc{FN17+SWiG} & 36.9 & 34.7 & 24.5 \\
        \textsc{ACE+SWiG\textsubscript{M2E2}} & 55.1 & 36.9 & 31.7 \\
        \midrule
        \multicolumn{4}{l}{\textbf{\textit{Pred Triggers}}} \\
        \textsc{FN17+SWiG} & 25.8 & 17.4 & 20.7 \\
        \textsc{ACE+SWiG\textsubscript{M2E2}} & 38.2 & 20.9 & 27.4 \\
        \bottomrule
    \end{tabular}}
    \caption{Evaluaton of FrameNet transfer learning. \textit{Gold Triggers} represent EAE based on annotated event mentions. \textit{Pred Triggers} represent EAE based on predicted event mentions.}
    \label{table:results:framenet}
\end{table}
\section{Conclusion}\label{sec:conclusion}

In this work, we investigate the multimedia EAE task and propose \textsc{MMUTF}, a simple yet effective unified template filling approach that assigns candidates to corresponding argument roles.
We encode entity candidates and object candidates from textual and visual inputs, respectively.
Then, we feed a natural language event template into the query decoder, which facilitates interactions with document contexts via cross-attention mechanisms.
Finally, candidate and query representations serve as the basis for computing candidate-query matching scores.
Our results demonstrate the effectiveness of the model and highlight the benefits of cross-ontology transfer and event semantic incorporation.
This assumption is supported by SRL transfer learning experiments, which we adapt to the multimedia EAE task.
Future work will focus on incorporating additional textual and visual cues from input documents, integrating event detection, and exploring bi-encoder architectures to reduce inference time.
\section*{Limitations}\label{sec:limitations}
In this work, we propose a unified template-filling approach for solving multimodal multimedia event argument extraction. 
Our method relies on manually crafted prompts that represent the event templates for each event type. 
The time required to construct such prompts was reasonable for the M2E2 benchmark, which covers only 8 event types and 15 argument roles for solving this task.
Other event extraction ontologies cover over 100 event types and complex argument relations \citep{huang_textee_2023}. 
Although we used a semi-automatic approach for the more comprehensive FrameNet, we still identified erroneous event templates, indicating that this method is not fully reliable.
The manual prompt construction for such large event ontologies also poses challenges in terms of required time and complexity. 
In addition, our system relies on event predictions, textual entities, and visual objects generated by preceding systems.
This approach is prone to error propagation (cf. \ref{sec:appendix:errors}) and requires a holistic assessment of all system components with appropriate evaluation data.

% Bibliography entries for the entire Anthology, followed by custom entries
%\bibliography{anthology,custom}
% Custom bibliography entries only
\bibliography{main}

\clearpage
\appendix

\section{Appendix}\label{sec:appendix}

% NOTE: Moved to introduction
% \subsection{Event Example}\label{sec:appendix:example}

% In \autoref{fig:example}, we depict a representative example from the M2E2 benchmark \citep{li_cross-media_2020}.
% Event argument roles such as \textit{\textbf{vehicle}} and \textit{\textbf{destination}} are only present in the visual and textual modality, respectively. 

% \begin{figure}[h]
%   \centering
% \includegraphics[width=1.0\linewidth]{examplev1.pdf}
%   \vspace{-8mm}
%   \caption{An example of Multimedia Event Extraction from the M2E2 benchmark, where event argument roles are extracted from both the textual and visual modality.}
%   \label{fig:example}
% \end{figure}

\subsection{Baselines}\label{sec:appendix:baselines}

We compare our proposed approach with a wide range of SOTA models.
1) WASE \citep{li_cross-media_2020} uses graph neural networks (GNN) together with abstract meaning representations and situation graphs for text and image, respectively.
These models additionally include the VOA image-caption dataset for weak cross-modality alignment, while the 2) FLAT \citep{li_cross-media_2020} baseline removes the GNN components.
3) CLIP\textsubscript{\textsc{Event}} \citep{li_clip-event_2022} combines the previous approaches with pre-trained CLIP model weights and introduces a Optimal Transport strategy for event structure representation and modality alignment.
4) UniCL \citep{liu_multimedia_2022} proposes a unified contrastive learning framework, uses a multimodal contrastive learning mechanism and incorporates textual and visual clues.
5) CAMEL \citep{du_training_2023} is based on cross-modal data augmentation, integrates an adapter module and applies a specialized training strategy.
6) MGIM \citep{liu_multi-grained_2024} is a multi-grained gradual inference model with fine-grained alignment between texts and images by multiple rounds of gradual inference.
7) UMIE \citep{sun_umie_2024} utilizes instruction tuning for unified multimodal information extraction based on text generation and Flan-T5 \citep{chung_flan_2024} in various model sizes.

\subsection{Threshold Tuning}\label{sec:appendix:threshold}

\begin{figure}[t]
  \centering
\includegraphics[width=1.0\linewidth]{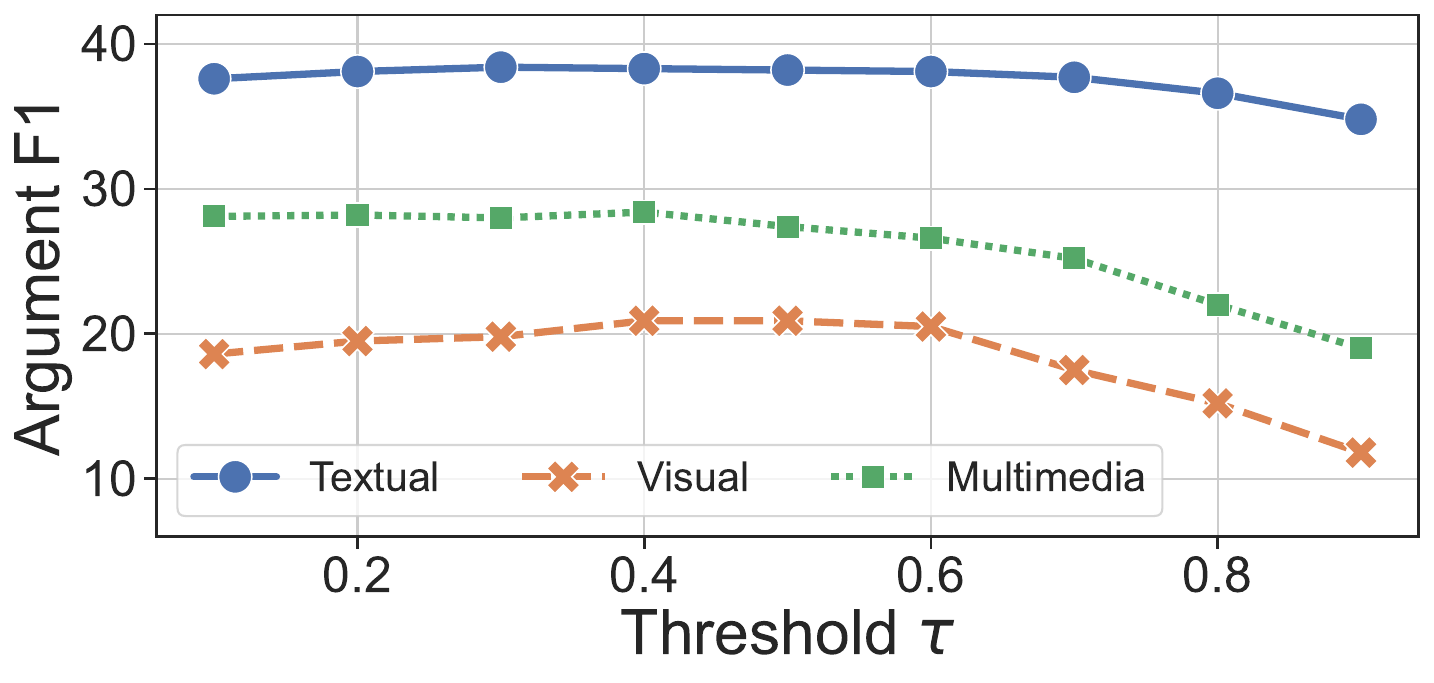}
  \vspace{-8mm}
  \caption{EAE F1 results for our \textsc{MMUTF} model with varying thresholds for each modality.}
  \label{fig:threshold}
\end{figure}

We investigate the influence of an automatic or manually chosen threshold for both the textual and visual matching scores.
In our work, we select the default threshold $\tau = 0.5$ and obtain reasonable results.
\autoref{fig:threshold} shows the metrics for different thresholds and modalities.
Additionally, we analyze threshold tuning based on the best F1-scores for the ACE and SWiG test datasets.
This procedure yields $\tau^{T} = 0.2$ and $\tau^{V} = 0.3$ for text and vision, respectively.
For automatic threshold tuning, we obtain a multimedia F1-score of 28.3\%, improving our default $\tau$ by +0.9\%.
In addition, we find the most performance degradation above a threshold of 0.6 while achieving the best results between 0.3 and 0.5.
This suggests a weak discriminative power between positive and negative candidate-query pairs.
Contrastive learning and hard negative mining may address this issue, but is not within the scope of this work.

\begin{table*}[t]
    \centering
    \resizebox{\linewidth}{!}{
    \begin{tabular}{ll}
    \toprule
        \textbf{Event Type} & \textbf{Event Template} \\
        \midrule
        \multicolumn{2}{l}{\textbf{\textit{M2E2}}} \\
        Movement:Transport & \textbf{\textit{[Agent]}} transported \textbf{\textit{[Artifact]}} in \textbf{\textit{[Vehicle]}} from \textbf{\textit{[Origin]}} to \textbf{\textit{[Destination]}}. \\
        Conflict:Attack & \textbf{\textit{[Attacker]}} attacked \textbf{\textit{[Target]}} using \textbf{\textit{[Instrument]}} as \textbf{\textit{[Place]}}. \\
         Conflict:Demonstrate & \textbf{\textit{[Entity]}} protested besides \textbf{\textit{[Police]}} using \textbf{\textit{[Instrument]}} at \textbf{\textit{[Place]}}. \\
         Justice:Arrest-Jail & \textbf{\textit{[Person]}} was sent to jail or arrested by \textbf{\textit{[Agent]}} using \textbf{\textit{[Instrument]}} at \textbf{\textit{[Place]}}. \\
         Contact:Phone-Write & \textbf{\textit{[Entity]}} called or texted messages using \textbf{\textit{[Instrument]}} at \textbf{\textit{[Place]}}. \\
         Contact:Meet & \textbf{\textit{[Entity]}} met at \textbf{\textit{[Place]}}. \\
         Life:Die & \textbf{\textit{[Agent]}} killed \textbf{\textit{[Victim]}} with \textbf{\textit{[Instrument]}} at \textbf{\textit{[Place]}}. \\
         Transaction:Transfer-Money & \textbf{\textit{[Giver]}} paid \textbf{\textit{[Recipient]}} with \textbf{\textit{[Money]}} for the benefit of \textbf{\textit{[Beneficiary]}} at \textbf{\textit{[Place]}}. \\
         \midrule
        \multicolumn{2}{l}{\textbf{\textit{FrameNet}}} \\
        Abandonment & \textbf{\textit{[Agent]}} abandon \textbf{\textit{[Theme]}} in \textbf{\textit{[Place]}}. \\
        Motion & \textbf{\textit{[Theme]}} move in \textbf{\textit{[Area]}} \textbf{\textit{[Direction]}} from \textbf{\textit{[from]}} to/into \textbf{\textit{[Goal]}} in \textbf{\textit{[Place]}}. \\
        Losing track of theme & \textbf{\textit{[Perceiver]}} lose \textbf{\textit{[Theme]}} in \textbf{\textit{[Place]}}. \\
        Adjusting & \textbf{\textit{[Agent]}} adjust \textbf{\textit{[Part]}} \textbf{\textit{[Feature]}} in \textbf{\textit{[Place]}} with \textbf{\textit{[Instrument]}}. \\
        Killing & \textbf{\textit{[Cause]}} \textbf{\textit{[Killer]}} kill \textbf{\textit{[Victim]}} with \textbf{\textit{[Instrument]}} \textbf{\textit{[Means]}} in \textbf{\textit{[Place]}}. \\
        Appointing & \textbf{\textit{[Selector]}} appoint to \textbf{\textit{[Body]}} \textbf{\textit{[Official]}} as \textbf{\textit{[Role]}} \textbf{\textit{[Function]}} in \textbf{\textit{[Place]}}. \\
        Arranging & \textbf{\textit{[Agent]}} arrange \textbf{\textit{[Theme]}} in \textbf{\textit{[Configuration]}} with \textbf{\textit{[Instrument]}}.\\
        Fining & \textbf{\textit{[Speaker]}} fine \textbf{\textit{[Payer]}} \textbf{\textit{[Fine]}} \textbf{\textit{[Reason]}} with \textbf{\textit{[Instrument]}} in \textbf{\textit{[Place]}}.\\
    \bottomrule
    \end{tabular}}
    \caption{M2E2 and FrameNet event templates for each event type serve as natural language prompts for EAE.}
    \label{tab:templates}
\end{table*}

\subsection{Implementation Details}\label{sec:appendix:implementation}

In our experiments, we use the implementation of the \textit{Transformers} \citep{wolf_2020} (v4.41.0) library in conjunction with PyTorch (v2.3.0).
For all runs, we evaluated the performance at each epoch on the ACE2005 test set and selected the best-performing checkpoint.
Unless otherwise mentioned, we use T5-base \citep{raffel_exploring_2020} with 222M parameters as text encoder-decoder and CLIP-base \citep{radford_learning_2021} with 85M parameters as vision encoder model.
All models are trained with mixed precision and executed on A100 GPUs with 40GB HBM using the same compute node running CUDA 12.3 and NVIDIA device driver version 545.
The average training time for each run did not exceed 3h.

\begin{table}[t]
    \centering
    \resizebox{\linewidth}{!}{
    \begin{tabular}{llccc}
        \toprule
        \multicolumn{2}{l}{\textbf{Model}} & \textbf{Textual F1} & \textbf{Visual F1} & \textbf{MM F1} \\
        \toprule
        \textsc{BERT} & \textsc{ViT} & 37.9 & 19.2 & 26.6 \\
         & \textsc{CLIP} & 37.6 & 19.7 & 26.6 \\
         & \textsc{Data2Vec} & 38.4 & 18.3 & 26.1 \\
        \midrule
        \textsc{BART} & \textsc{ViT} & 38.1 & 19.2 & 24.9 \\
         & \textsc{CLIP} & 38.5 & 20.4 & 27.9 \\
         & \textsc{Data2Vec} & 39.0 & 18.4 & 27.0 \\
        \midrule
        \textsc{T5} & \textsc{ViT} & 38.3 & 19.5 & 27.6 \\
         & \textsc{CLIP} & 38.2 & 20.9 & 27.4 \\
         & \textsc{Data2Vec} & 38.5 & 18.5 & 27.0 \\
        \bottomrule
    \end{tabular}}
    \caption{Results for text and vision models.}
    \label{table:appendix:backbones}
\end{table}

\subsection{Event Templates}\label{sec:appendix:templates}

All manually crafted event templates for the M2E2 benchmark are shown in \autoref{tab:templates}.
The words in brackets correspond to the argument roles we aim to extract.
These event templates are selected based on the predicted event type, then tokenized and fed into the query model to obtain the final query representations.
For FrameNet, it is infeasible to manually construct templates for each frame and corresponding frame element.
Therefore, we follow \citep{zhang_transfer_2022} and adopt their semi-automatic method to generate templates for our transfer learning experiments.
We merge the ontologies of FrameNet 1.7 \citep{baker_berkeley_1998} and SWiG \citep{vedaldi_grounded_2020} and use the FrameNet templates for both datasets.
We show some excerpts in \autoref{tab:templates}.

\subsection{Backbone Models}\label{sec:appendix:backbones}

We compare different text and vision model configurations in \autoref{table:appendix:backbones}.
For the pre-trained text models, we compare the following encoder and encoder-decoder variants: \textsc{BERT}\footnote{https://huggingface.co/google-bert/bert-base-uncased} \citep{devlin_bert_2019}, \textsc{BART}\footnote{https://huggingface.co/facebook/bart-base} \citep{lewis_bart_2020}, and \textsc{T5}\footnote{https://huggingface.co/google-t5/t5-base} \citep{raffel_exploring_2020}.
Regarding \textsc{BERT}, we use two distinct models for candidate and query representations and add cross-attention to the query encoder.
For the pre-trained vision models, we rely on \textsc{ViT}\footnote{https://huggingface.co/google/vit-base-patch16-224} \citep{dosovitskiy_2021}, \textsc{CLIP}\footnote{https://huggingface.co/openai/clip-vit-base-patch16} \citep{radford_learning_2021}, and \textsc{Data2Vec}\footnote{https://huggingface.co/facebook/data2vec-vision-base} \citep{baevski_2022}.
All models represent their corresponding base variants, as we did not observe substantial improvements with larger model sizes.

Overall, the results demonstrate robust performance across different model configurations. 
However, we observe that \textsc{CLIP} generally outperforms the other vision models in terms of visual and multimedia F1 scores.
We hypothesize that pre-training on large image-caption datasets is beneficial for the M2E2 task.

\begin{table}[t]
    \centering
    \resizebox{\linewidth}{!}{
    \begin{tabular}{lccc}
        \toprule
        \textbf{Template} & \textbf{Textual F1} & \textbf{Visual F1} & \textbf{MM F1} \\
        \toprule
        \multicolumn{4}{l}{\textbf{\textit{w/ event type}}} \\
        Concatenation & 38.0 & 18.9 & 27.9 \\
        Standard & 38.3 & 17.9 & 25.3 \\
        Enriched & 38.5 & 16.8 & 25.2 \\
        \midrule
        \multicolumn{4}{l}{\textbf{\textit{w/o event type}}} \\
        Concatenation & 38.7 & 19.0 & 28.0 \\
        Standard & 38.2 & 20.9 & 27.4 \\
        Enriched & 38.0 & 19.7 & 28.1 \\
        \bottomrule
    \end{tabular}}
    \caption{Results for different prompt templates.}
    \label{table:appendix:prompts}
\end{table}

\subsection{Prompt Analysis}\label{sec:appendix:prompts}

In \autoref{table:appendix:prompts}, we investigate prompt template variations and conduct experiments with and without event type prefixes. 
For event type prefixes, we prepend the event type name to the event template (e.g., "Conflict Attack: <Event Template>").
We compare three different prompt variations: 
\textit{Concatenation} simply concatenates the argument role names from the event templates. \textit{Standard} represents the manually crafted event templates used in this work (cf. \autoref{tab:templates}).
\textit{Enriched} adds argument role definitions after each argument role name.

The results show that event templates without event type prefixes generally outperform their counterparts across all variations.
We conjecture this is because similar arguments appear in multiple events, and event type prefixes may impede the transfer of knowledge between roles.
All variations perform similarly for textual EAE, while event type prefixes slightly impair model performance for visual and multimedia EAE.

\subsection{Error Analysis}\label{sec:appendix:errors}

While our method achieves significant improvements across all SOTA models for the textual EAE task, we still recognize shortcomings for the visual modality.
Therefore, we analyse the impact of visual event mentions and visual candidates predicted by YOLOv8 \citep{jocher_ultralytics_2023}.
With gold triggers (i.e., annotated visual event mentions), we achieve an absolute improvement of 16\% in visual F1 which highlights the importance of reliable ED.
However, the inclusion of gold candidates (i.e., annotated bounding boxes) demonstrates an absolute improvement of 35.5\% visual F1 which further enhances the gold triggers by 19.5\%.
These results suggests that the correct selection of candidates is crucial and the joint modeling of candidate generation and EAE poses a promising direction.

\end{document}